\def\year{2021}\relax
\documentclass[11pt]{article}
\usepackage{naacl2021}
\usepackage[T1]{fontenc}

\usepackage[utf8]{inputenc}
\usepackage{microtype}
\usepackage{tabularx}
\usepackage{times}  
\usepackage{helvet} 
\usepackage{amssymb} 
\usepackage{courier}  
\usepackage{booktabs, multirow} 
\usepackage{soul}
\usepackage{changepage,threeparttable} 
\usepackage{graphicx} 
\usepackage{natbib}  
\usepackage[belowskip=0pt,aboveskip=0pt]{caption,subcaption} 
\usepackage{comment}
\usepackage{amsmath}

\usepackage{xcolor}
\usepackage{amsfonts,amssymb}
\usepackage[ruled,vlined,linesnumbered]{algorithm2e} 
\usepackage{enumitem}

\newcommand{\comments}{0} 
\ifthenelse{\equal{\comments}{1}}
{

\newcommand{\thua}[1]{{\color{blue}{}}}
}

\usepackage{latexsym}

\SetKwInput{KwInput}{Input}                
\SetKwInput{KwOutput}{Output}              
\SetKwInput{KwRequire}{Require}              
\usepackage{microtype}
\usepackage{comment}
\usepackage{setspace} 
\usepackage{enumitem}
\usepackage{float}
\restylefloat{table}
\usepackage{lipsum}
\usepackage{multirow}

\newenvironment{packed_item}{
	\begin{itemize}[leftmargin=*,topsep=0pt,partopsep=0pt]
		\setlength{\itemsep}{0pt}
		\setlength{\parskip}{0pt}
		\setlength{\parsep}{0pt}
	}
	{\end{itemize}}

\newenvironment{packed_enum}{
	\begin{enumerate}[leftmargin=*,topsep=0pt,partopsep=0pt]
		\setlength{\itemsep}{0pt}
		\setlength{\parskip}{0pt}
		\setlength{\parsep}{0pt}
		\setlength{\topsep}{0pt}
	}
	{\end{enumerate}}

\title{Hyperparameter-free Continuous Learning for Domain Classification in Natural Language Understanding}
\author{Ting Hua \And
  Yilin Shen \And
  Changsheng Zhao \\
    Samsung Research America \\
    \texttt{\{ting.hua,yilin.shen,changsheng.z,yenchang.hsu,hongxia.jin\}@samsung.com}
  \And
  Yen-Chang Hsu\And
  Hongxia Jin

  }

\begin{document}
\maketitle

\begin{abstract}
Domain classification is the fundamental task in natural language understanding (NLU),  which often requires fast accommodation to new emerging domains.
This constraint makes it impossible to retrain all previous domains, even if they are accessible to the new model. 
Most existing continual learning approaches suffer from low accuracy and performance fluctuation, especially when the distributions of old and new data  are significantly different.
In fact,  the key real-world problem is not the absence of old data, but the inefficiency to retrain the model with the whole old dataset.
Is it potential to utilize some old data to yield high accuracy and maintain stable performance, while at the same time, without introducing extra hyperparameters?
In this paper, we proposed a hyperparameter-free continual learning model for text data that can stably produce high performance under various environments. 
Specifically,  we utilize Fisher information to select exemplars that can ``record''  key information of the original model. 
Also, a novel scheme called dynamical weight consolidation  is proposed to enable  hyperparameter-free learning during the retrain process.
Extensive experiments demonstrate that baselines suffer from fluctuated performance and therefore useless in practice. 
On the contrary, our proposed model CCFI significantly and consistently outperforms the best state-of-the-art method by up to 20\% in average accuracy, and each   component of CCFI contributes effectively to overall performance.
\end{abstract}

\section{Introduction}

\textit{Catastrophic forgetting}  is the well-known Achilles' heel of deep neural networks, that the knowledge learned from previous tasks will be forgotten when the networks are retrained to adapt to new tasks.
Although this phenomenon has been noticed as early as the birth of neural networks \cite{french1999catastrophic, mccloskey1989catastrophic},
it didn't attract much attention until deep neural networks have achieved impressive  performance gains in  various applications \cite{lecun2015deep,krizhevsky2012imagenet}.

\textit{Domain classification} is the task that mapping the spoken utterances to natural language understanding domains.
It is widely used in intelligent personal digital assistant (IPDA) devices, such as Amazon Alexa, Google Assistant, and Microsoft Cortana.
As many IPDAs now allow third-party developers to build and integrate new domains \cite{kumar2017just}, these devices are eager to continual learning technologies that can achieve high performance stably \cite{Amazon2018Shorter_extension,Amazon2018Shorter}. 
However, most traditional IPADs only work with well-separated domains built by specialists \cite{tur2011spoken} or customized designed for specific datasets \cite{Amazon2019continuous}.

There is still a lack of continual learning methods that capable of general domain classification. Most previous approaches capable of continual learning focus on the scenario that the new model should be retrained without any access to old data \cite{Lwf2016learning,EWC2017,GEM2017gradient}.
However, these methods often involve more parameters that require extensive efforts in expert tuning. And when data distributions of new tasks  are obviously different from the original task (e.g. class-incremental learning), these approaches can hardly maintain good accuracy for both tasks and may suffer from fluctuations in performance. 
On the other hand, old data are not unavailable in many practical cases.
The main concerns arise from the tremendous cost in memory and computation resources, if the model is updated with huge previous datasets. 
Also, most continual learning approaches are applied to image data that little attention has been paid to text data.
Is it possible to develop a desirable model capable of continual learning that satisfies the following qualities?
1) \textbf{High accuracy with limited old data}. 
Compared to the extreme cases that no access or full access to old data, it is more practical to put models under the setting with a limited amount of old data available (e.g., no more than 10\% of original data).
In this case, models can achieve good performance without too much additional cost in physical resources and can be conveniently renewed with periodical system updates.
2) \textbf{High stability  with zero extra parameters}.
Many continual learning models can perform well only under restricted experiment settings, such as specific datasets or carefully chosen parameters. 
However, practical datasets are usually noisy and imbalanced distributed, and inexperienced users can't set suitable parameters in real-world applications. Therefore, it is desirable to develop a hyperparameter-free model that can work stably under various experimental environments.

To achieve these goals, we proposed a \emph{{C}ontinous learning model based on weight {C}onsolidation and {F}isher {I}nformation sampling} (CCFI), with application to domain classification.
The main challenge is how to ``remember'' information from original tasks, not only the representative features from data, but also the learned parameters of the model itself. This is a non-trivial contribution since ``uncontrollable changes'' will happen to neural network parameters whenever the feature changes. 
To avoid such ``uncontrollable changes'', previous work iCarL even discards deep neural networks as its final classifier and turns to k-nearest neighbors algorithm for actual prediction \cite{icarl2017}.
Our work demonstrates that these changes are ``controllable'' with exemplar selected by Fisher information and parameters learned by Dynamical Weight Consolidation. 
Our contributions can be summarized as follows. 
\begin{packed_item}
    \item \textbf{Fisher information sampling}. Good exemplars are required to ``remember'' key information of old tasks. Unlike previous work using simple mean vectors to remember the information of old data, exemplars selected by Fisher information record both the features of data and the information of the original neural network. 
    \item \textbf{Dynamical weight consolidation}. The need for hyperparameter is an inherited problem of regularization-based continual learning. Previous works search for this hyperparameter by evaluating the whole task sequence, which is supposed not to be known. This work provides a simple auto-adjusted weighting mechanism, making the regularization strategy possible for a practical application.
    Also, traditional weight consolidation methods such as EWC \cite{EWC2017} are designed for sequential tasks with similar distributions. We extend it to the incremental learning scenario and add more regularity to achieve better stability. 
    \item \textbf{Extensive experimental validation}. 
    Most existing continual learning methods are designed for image data, while few previous attempts working on text data are often limited to specific usage scenarios and rely on fine-tuned parameters. 
    Our proposed CCFI model is a general framework that can be efficiently applied to various environments with the least efforts in parameter tuning.
    Our extensive experimental results demonstrate the proposed CCFI can outperform all state-of-the-art methods, and provide insights into the working mechanism of methods. 
\end{packed_item}


\section{Related Work}
\label{sec:relatedwork} 


Although most of the existing approaches are not directly applicable to our problem, several main branches of research related to this work can be found:  exemplar selection,  regularization of weights, and feature extraction or fine-tune method based on pre-trained models.


Our problem is closest to the setting of exemplar selection methods \cite{icarl2017,Amazon2019continuous}.
These approaches store examples from original tasks, and then combine them together with the new domain data for retraining.
 iCarL \cite{icarl2017} discards the classifier based on neural network to prevent the catastrophic forgetting, and turns to traditional K-nearest neighbor classifier.

To avoid large changes of important parameters,
regularization models \cite{EWC2017,Lwf2016learning,zenke2017continual} add constraints to the loss function. They usually introduce extra parameters requiring careful initialization.  And  it has been shown that their performance will drop significantly if the new tasks are drawn from different distributions \cite{icarl2017}. 
On the contrary,  our proposed CCFI is a parameter-free model that can produce stable performance under various experimental environments. 

Feature extraction methods utilize pre-trained neural networks to calculate features of input data \cite{donahue2014decaf,sharif2014cnn}.
They make little modifications to the original network but often result in a limited capacity for learning new tasks. 
Fine-tuning models \cite{girshick2014rich} can modify all the parameters in order to achieve better performance in new tasks. Although starting with a small learning rate can indirectly preserve the knowledge learned from the original task, fine-tuning method will eventually tend to new tasks. 
Adapter tuning \cite{houlsby2019parameter} can be viewed as the hybrid of fine-tune and feature extraction.
Unlike our model that makes no changes to the backbone model, the Adapter tuning method increases the original model size and introduces more parameters by inserting adapter modules to each layer.

\section{Our CCFI Model}
\label{sec:model}
Given data stream  $D=\{x_i,y_i\}_{i=1}^N$, 
the  classification tasks in deep learning neural networks are equal to find the best parameter set $\Theta$ that can maximize the probability of the data $p(\mathcal{D}|\Theta)$.
Namely, the classifier can  make predictions $\hat{Y}$ that best reproduce the ground truth labels $Y$.
Under the continual learning setting, new data $\mathcal{D}^n$ of additional classes will be added to the original data stream $\mathcal{D}^o$ in an incremental form. 
Our goal is to update the old parameters $\Theta^o$ (trained on original data stream $\mathcal{D}^o$) to the new parameters $\Theta^{n}$, which can work well on both new data $\mathcal{D}^n$ and old data $\mathcal{D}^o$.

In this paper, the initial model is trained with the original data set $\mathcal{D}^o$, and will output the trained model $\Theta^o$. In this training process, \textbf{Fisher Information Sampling}  is conducted to select the most representative examples that can help to ``remember'' the parameters of the initial trained model. 
In the retraining process, the renewed model is learned based on  \textbf{Dynamical Weight Consolidation}, and evaluated on the training set consisted of new classes and the old exemplars.

\subsection{Fisher Information Sampling}
\label{Fisher_info}
The critical problem of exemplar set selection is: what are good examples that can ``maintain'' the performance of old tasks?
The state-of-the-art method iCaRL \cite{icarl2017} selects examples close to mean feature representation, and CoNDA \cite{Amazon2019continuous} follows the same scheme to domain adaptation on text data.
To utilize the advantage of the mean feature and avoid catastrophic forgetting, iCaRL chooses k-nearest neighbors algorithm as the classifier rather than deep neural networks, although the latter is proved to be a much better performer.
Is there any exemplar selection method that can utilize the powerful deep learning models as the classifier, and at the same time, ``remember'' the key information of old tasks?

Fisher information measures  how much information that an observable random variable carries about the parameter.
For a parameter $\theta$ in the network $\Theta$ trained by data $\mathcal{D}$, its Fisher information is defined as the variance of the gradient of the log-likelihood:
\begin{eqnarray}
    && I(\theta) = Var \left(s(\theta)\right) \\
    &=& E{\left[\left( \frac{\partial }{{\partial \theta }}\log p(\mathcal{D}|\theta ) \right) \left( \frac{\partial }{{\partial \theta }}\log p(\mathcal{D}|\theta ) \right)^T \right]}.  \nonumber
    \label{equ:fisher_definition}
\end{eqnarray}
Fisher information can be calculated directly, if the exact form of $\log p(\mathcal{D}|\theta )$ is known.
However, the likelihood is often intractable. 
Instead, empirical Fisher information $\hat{I}(\theta)$
is computed through data $d_i \in \mathcal{D}$ drawn from $p(\mathcal{D}|\theta)$:
\begin{equation}
    \hat{I}(\theta) =\frac{1}{N}\sum\limits_{i = 1}^N {\left( {\frac{\partial }{{\partial \theta }}\log p({d_i}|\theta )} \right)} {\left( {\frac{\partial }{{\partial \theta }}\log p({d_i}|\theta )} \right)^T}.
    \label{equ:empirical_fisher}
\end{equation}

From another point of view, when $\log p(\mathcal{D}|\theta )$ is twice differentiable with respect to $\theta$, Fisher information can be written as the second derivative of likelihood: 
\begin{equation}
    I(\theta) = - E\left[ {\frac{{{\partial ^2}}}{{\partial {\theta ^2}}}\log p(\mathcal{D}|\theta )} \right].
    \label{equ:second-order}
\end{equation}
According to Equation \ref{equ:second-order}, three equivalent indications can be made to a high value in Fisher information $I(\theta)$:
\begin{packed_item}
    \item a sharp peak of likelihood with respect to $\theta$, 
    \item $\theta$ can be easily inferred from data $\mathcal{D}$,
    \item data $\mathcal{D}$ can provide sufficient information about the correct value for parameter $\theta$.
\end{packed_item}
Jointly thinking about the calculation form introduced by empirical Fisher information (Equation \ref{equ:empirical_fisher}) and the physical meaning of Fisher information revealed by the second derivative form (Equation \ref{equ:second-order}), 
we can find a way to measure how much information each data $d_i$ carries to the estimation of parameter $\theta$, which we call as empirical Fisher information difference:
\begin{equation}
    \Delta \hat{I}_i(\theta) = {\left( {\frac{\partial }{{\partial \theta }}\log p({d_i}|\theta )} \right)} {\left( {\frac{\partial }{{\partial \theta }}\log p({d_i}|\theta )} \right)^T}.
    \label{equ:Fisher_sample}
\end{equation}
Instead of simple mean feature vectors used in previous work \cite{icarl2017,Amazon2019continuous}, we use the empirical Fisher information difference to select exemplar set.
Specifically,  CCFI model makes use of BERT \cite{devlin2019bert} for text classification. 
 The base BERT model is treated as feature extractor $\Phi: X\rightarrow H$, which takes input token sequences $X$, and outputs the hidden representations $H$. 
To predict the true label $Y$, a softmax classifier is added to the top of BERT:
\begin{equation}
    p(\hat{Y}|X,\Theta)=\sigma(W^T \cdot H)= \sigma(W^T \cdot \Phi(X)),
    \label{equ:likelihood}
\end{equation}
where $W$ is the  task-specific parameter matrix for classification. 
The trained parameters $\Theta$ can therefore be split into the fixed feature extraction part $\Phi$ and variable weight parameter $W$.
In continual learning setting, we  denote $W^k \in {R^{h \times k}} $ as the most up-to-date weight matrix, where $k$ is the number of classes that have been so far observed, and $h$ is the size of the final hidden state $H$.

Remember that, for the classification task, the best parameters that can  maximize the probability of the data $p(\mathcal{D}|\Theta)$ are also the parameters that make predictions $\hat{Y}$ closest to the ground truth label $Y$. 
Therefore, we can take Equation \ref{equ:likelihood} into Equation \ref{equ:Fisher_sample}, 
in order to get the practical computation of empirical Fisher information difference for data $d_i$ on parameter $\theta$.
Since the parameters of feature extractor $\Phi$ are fixed,
only empirical Fisher information difference
of parameters in weight matrix ${w_j} \in W$ are calculated:
\begin{equation}
\Delta {\hat I_i}({{w_j}}) = \left( {\frac{\partial }{{\partial {{w_j}}}}\log {{[\sigma ({W^T} \cdot \Phi ({x_i}))]}^{({y_i})}}} \right)^2,
\label{equ:sampling_w}
\end{equation}
where the likelihood $p(d_i|\theta)$ is calculated via the log-probability value of the correct label $y_i$ of input $x_i$.
And the total empirical Fisher information difference data $d_i$ carrying is the sum over all $w_{j} \in W$:
\begin{equation}
\Delta {\hat I_i} = \sum\limits_{j = 1}^{h \times k} {\Delta {{\hat I}_i}({w_j})}.
\label{equ:sampling_i}
\end{equation}
Algorithm \ref{alg:exemplar} describes the exemplar selecting process. 
Within each class $k$, the samples top ranked by empirical Fisher information difference are selected as exemplars, till the targeted sample rate (e.g., $1\%$) is met. 
\begin{algorithm}[t]\small
\DontPrintSemicolon
  \KwInput{original data stream $\mathcal{D}^o$}
  \KwInput{trained neural network $\Theta^o= \{\Phi, W^o\}$}
  \KwInput{sample rate $r$}
  \For{each data  $d_i$}
  {
    \For{each parameter $w_j \in W^o$}
    { calculate ${\Delta {{\hat I}_i}({w_j})}$ using Equation \ref{equ:sampling_w}}
    calculate ${\Delta {{\hat I}_i}}$ using Equation \ref{equ:sampling_i}
  }
  \For{each class $k$}
  {
  rank the samples $d_i$ by  ${\Delta {{\hat I}_i}}$\;
  select the top $|\mathcal{D}_k|\times r$\ examples  as $E_k$
  }
  \KwOutput{exemplar set $E\leftarrow \{E_1,...,E_k\}$}
\caption{Construction of exemplar set}
\label{alg:exemplar}
\end{algorithm}

\begin{figure*}[htbp]
  \begin{subfigure}[b]{.33\textwidth}
   \centering
   \includegraphics[width=\columnwidth]{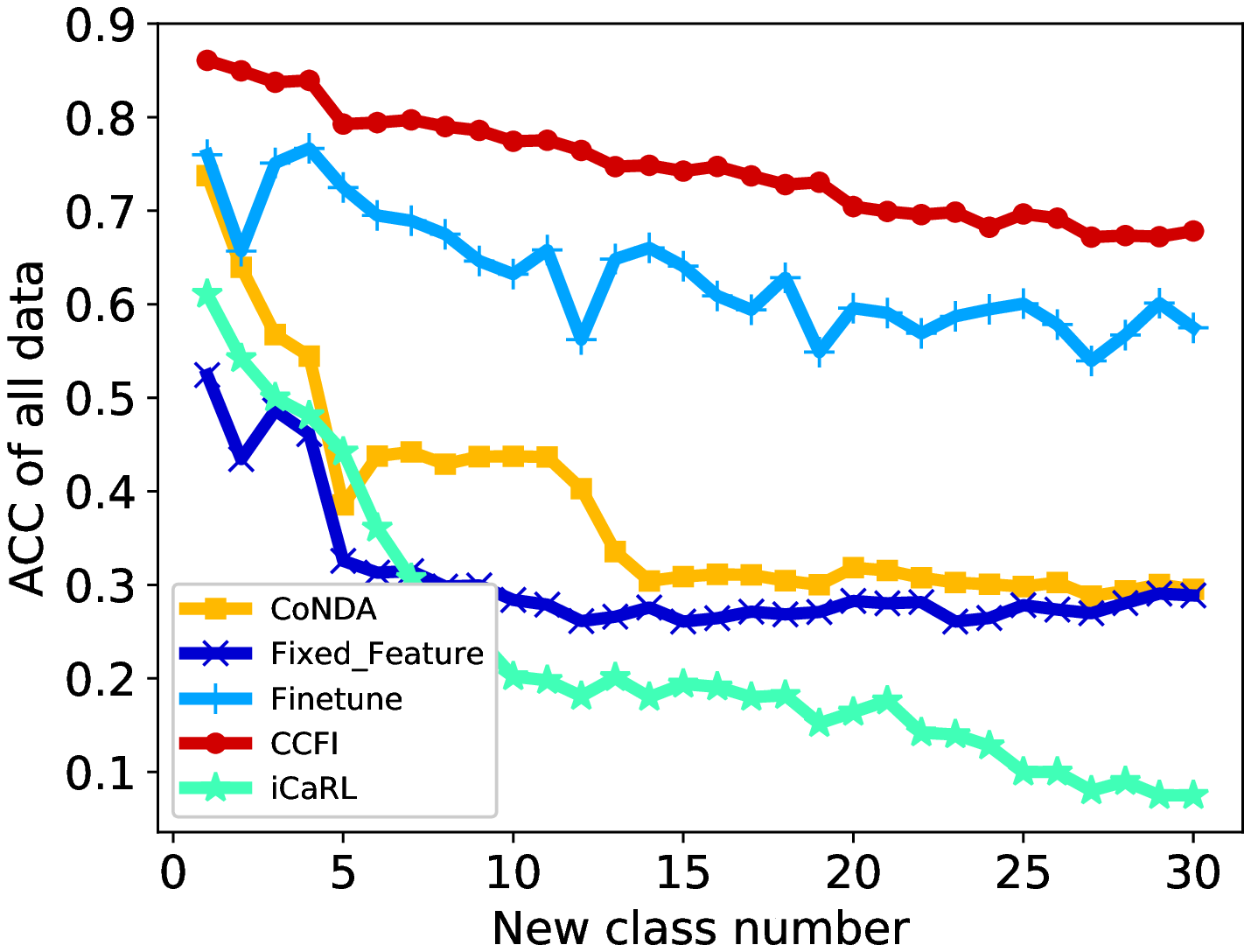}
       \caption{
      Overall accuracy. }
       \label{fig:cc_add_class_73:all}
       \end{subfigure}
   \begin{subfigure}[b]{.33\textwidth}
\includegraphics[width=\columnwidth]{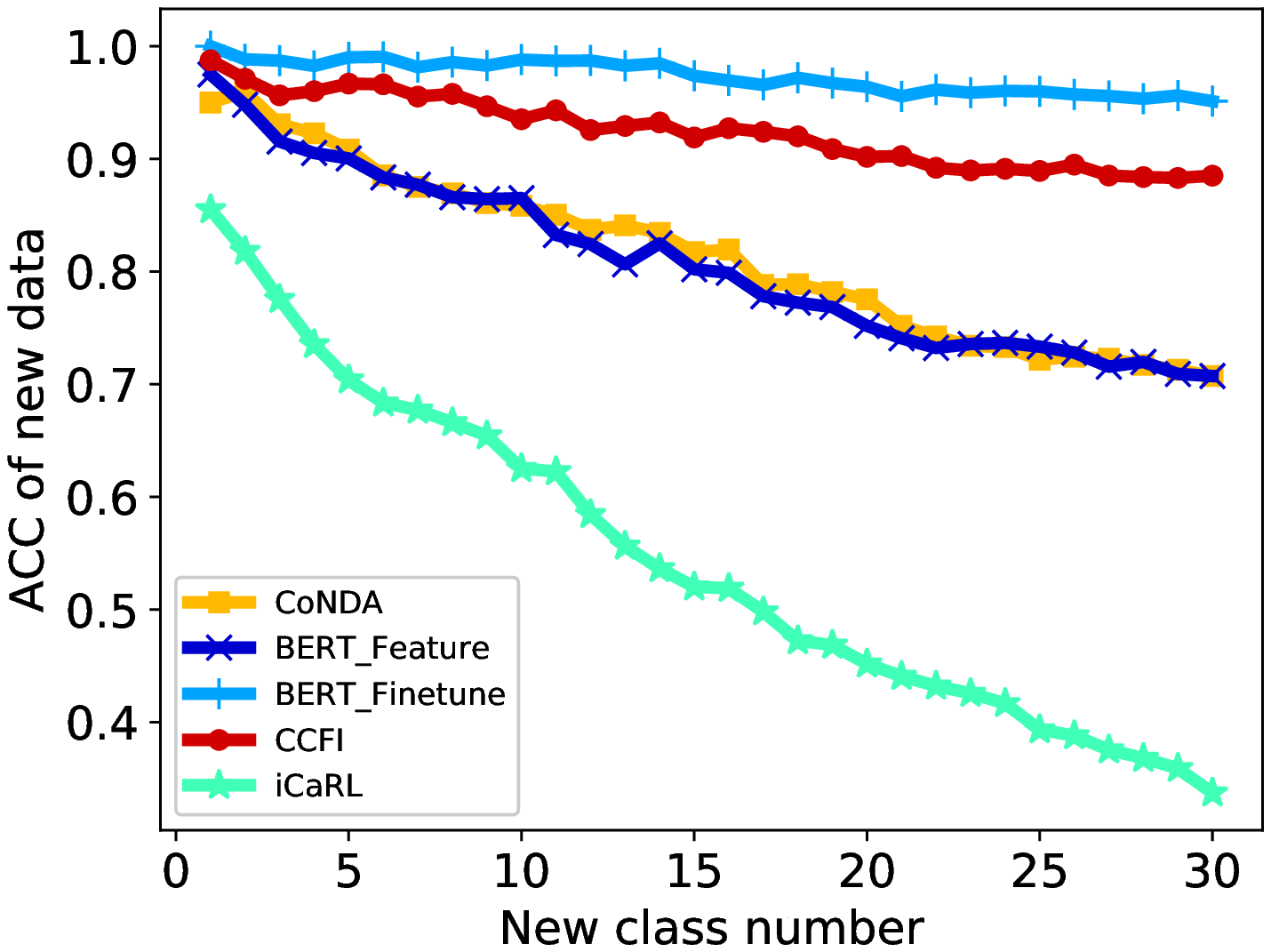}
\centering
  \caption{New task accuracy.}
   \label{fig:cc_add_class_73:new}
\end{subfigure}
   \begin{subfigure}[b]{.33\textwidth}
\includegraphics[width=\columnwidth]{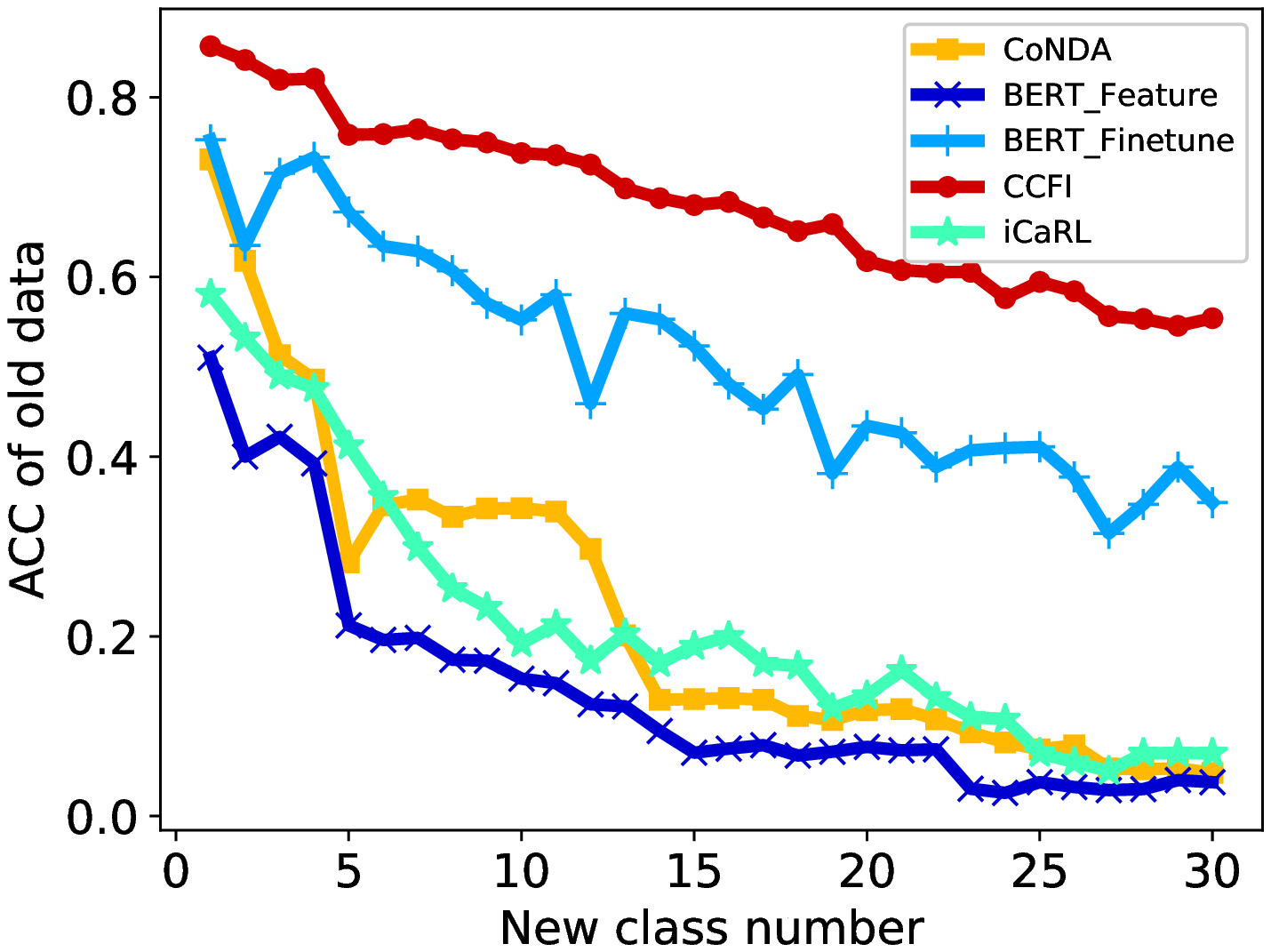}
\centering
  \caption{Old task accuracy.}
   \label{fig:cc_add_class_73:old}
   \end{subfigure}
 \caption{Performance of class-incremental learning on the 73-domain dataset. 
The observable ratio of old data is kept as 1\%, 43 out of 73 domains are used for initial training, and then 30 domains are added one-by-one for retraining. }
\label{fig:cc_add_class_73}
\end{figure*}

\subsection{Dynamical Weight Consolidation}
\label{sec:ewc}
The main goal of retraining process is: how to achieve good performance for both new and old tasks? 
EWC \cite{EWC2017} is proved to be a good performer that can balance the performance of old and new task.
However, rather than incremental learning problem studied in this paper, EWC is designed for the tasks with same class number but different in data distributions.
Furthermore, EWC requires careful hyperparameter setting, which is unrealistic to be conducted by inexperienced users.
In this section, we introduce a scheme named Dynamical Weight Consolidation, which can avoid the requirement of such hyperparameter setting.
Also, this scheme is demonstrated to perform more stably than traditional EWC in the experimental part.

Specifically, our loss function during the retraining process can be viewed the sum of two parts:
loss $\ell_n$ calculated by the correct class indicators of new data and loss $\ell_e$ to reproduce the performance of old model:
\begin{subequations}
	\begin{align}
		\label{equ:eq1}
 		& \ell_{n}= -\sum\limits_{y \in  {Y^n} } {y\log \hat y}, \\
 		\label{equ:eq2}
 		& \ell_{e}= -\sum\limits_{y \in {E} } {y\log \hat y} + \frac{\lambda }{2}\sum\limits_{j = 1}^{h \times (k_o+k_n)} {\hat I(w_j){{(\hat w_j^n - w_j^o)}^2}}.
	\end{align}
\end{subequations}
The loss function $\ell_e$ can be further divided into two parts:
the cross entropy of exemplar set, and the consolidation loss caused by modifying parameters with high Fisher information.
In traditional EWC model, the weight $\lambda$ that balances cross entropy and consolidation loss is a fixed value.
In our CCFI model, $\lambda$ is updated dynamically based on current values of cross entropy and consolidation loss:
\begin{equation}
    \lambda=\left\lfloor {\lg \frac{{ - \sum\limits_{y \in  {Y^n} } {y\log \hat y} }}{{\sum\limits_{j = 1}^{h \times (k_o+k_n)} {\hat I(w_j){{(\hat w_j^n - w_j^o)}^2}} }}} \right\rfloor.
    \label{equ:lambda}
\end{equation}
Note that, the $\hat I(w_j)$ is the element in the updated  parameter  information  table $T^n$.  The details can be found in the section \ref{ph:retraining}.


\subsection{Overall Process}
This part describes the overall process of the CCFI model. 
First we list the outputs of the old tasks,
then we introduce the detailed implements of retraining.

\subsubsection{Initial training}
After training the model with  old data ($k_o$ classes), the outputs of the old task include:1) trained model $\Theta^o$; 2) exemplars $E$ of old data, and 3) parameter information table $T^o$.
Each element in the parameter information table $T^o$ is the  empirical Fisher information $\hat I(w_j^o)$ of the old task, which can be computed through Equation \ref{equ:empirical_fisher} during the initial training  process.

\subsubsection{Retraining}
\label{ph:retraining}
The retraining process can be described as follows:
\begin{packed_enum}
    \item \textbf{Load freeze feature extractor}:  The feature extractor $\Phi$ is kept unchanged, which means the BERT encoder with transformer blocks and  self-attention heads are freezed. 
    \item \textbf{Update variable weight matrix}:  To adopt the new classes data $X^n$, the original variable weight matrix $W^{k_o}$ is extended to $W^{k_o+k_n} \in R^{h \times (k_o+k_n)}$, where the first $k_o$ columns are kept the same with  the original model and the new $k_n$ columns are initialized with random numbers.
    \item \textbf{Update parameter information table}:  Similar to  variable  weight  matrix, parameter information table $T^o$ is a matrix with dimension ${h \times k_o}$.
    To adopt the new data, it is extended to the new matrix $T^n$ with dimension ${h \times (k_o+k_n)}$, where the first $k_o$ columns are same with  $T^o$ and the new $k_n$ columns are initialized with zero.
    In this way, the new model can freely update the the new $k_n$ columns to lower classification loss, but will receive penalty when modifying important parameters in the original $k_o$ columns. 
\end{packed_enum}

\begin{figure*}[htbp]
     \begin{subfigure}[b]{.33\textwidth}
   \centering
   \includegraphics[width=\columnwidth]{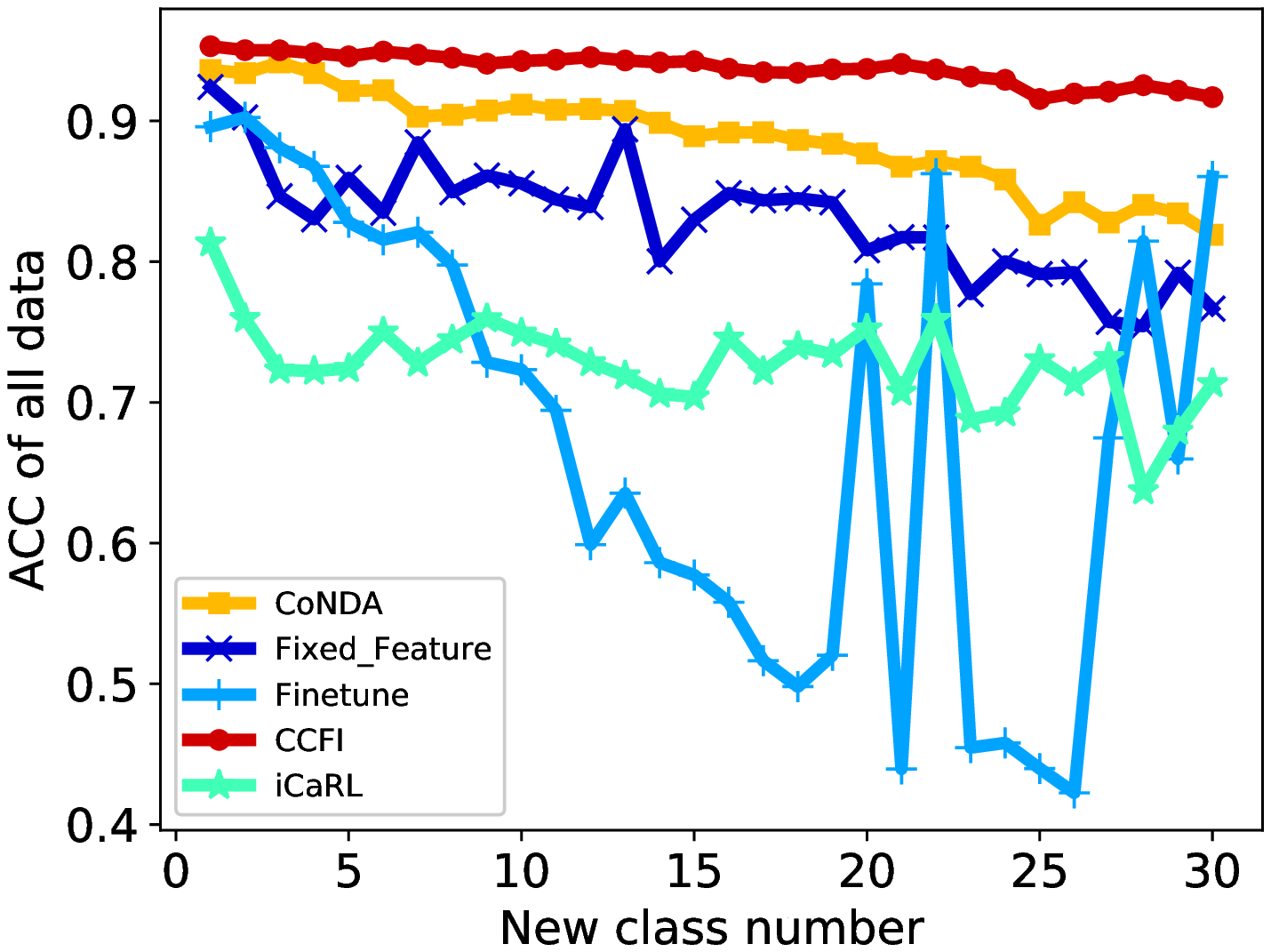}
       \caption{Overall accuracy. }
       \label{fig:cc_add_class_150:all}
       \end{subfigure}
   \begin{subfigure}[b]{.33\textwidth}
\includegraphics[width=\columnwidth]{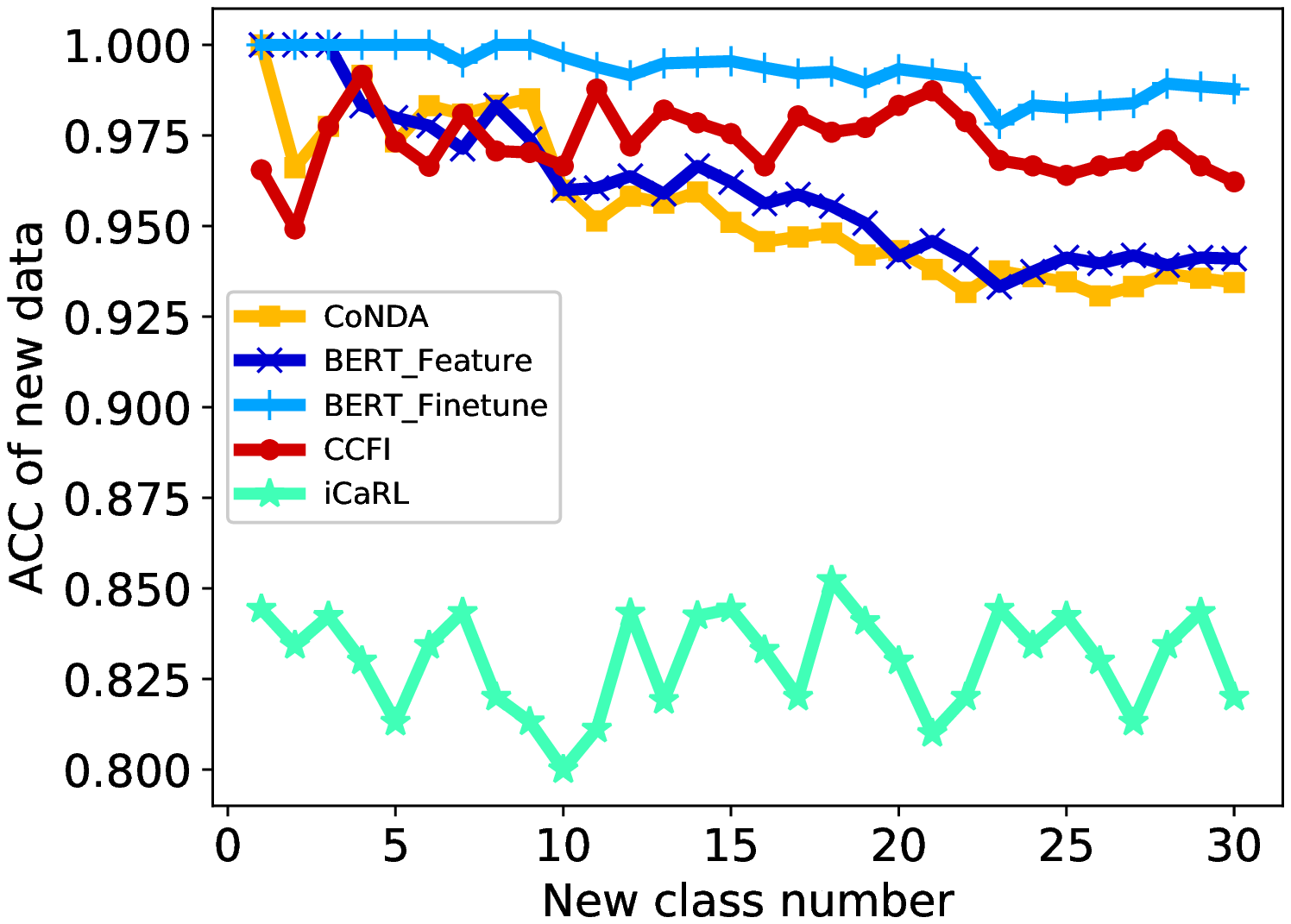}
\centering
  \caption{New task accuracy.}
   \label{fig:cc_add_class_150:new}
\end{subfigure}
   \begin{subfigure}[b]{.33\textwidth}
\includegraphics[width=\columnwidth]{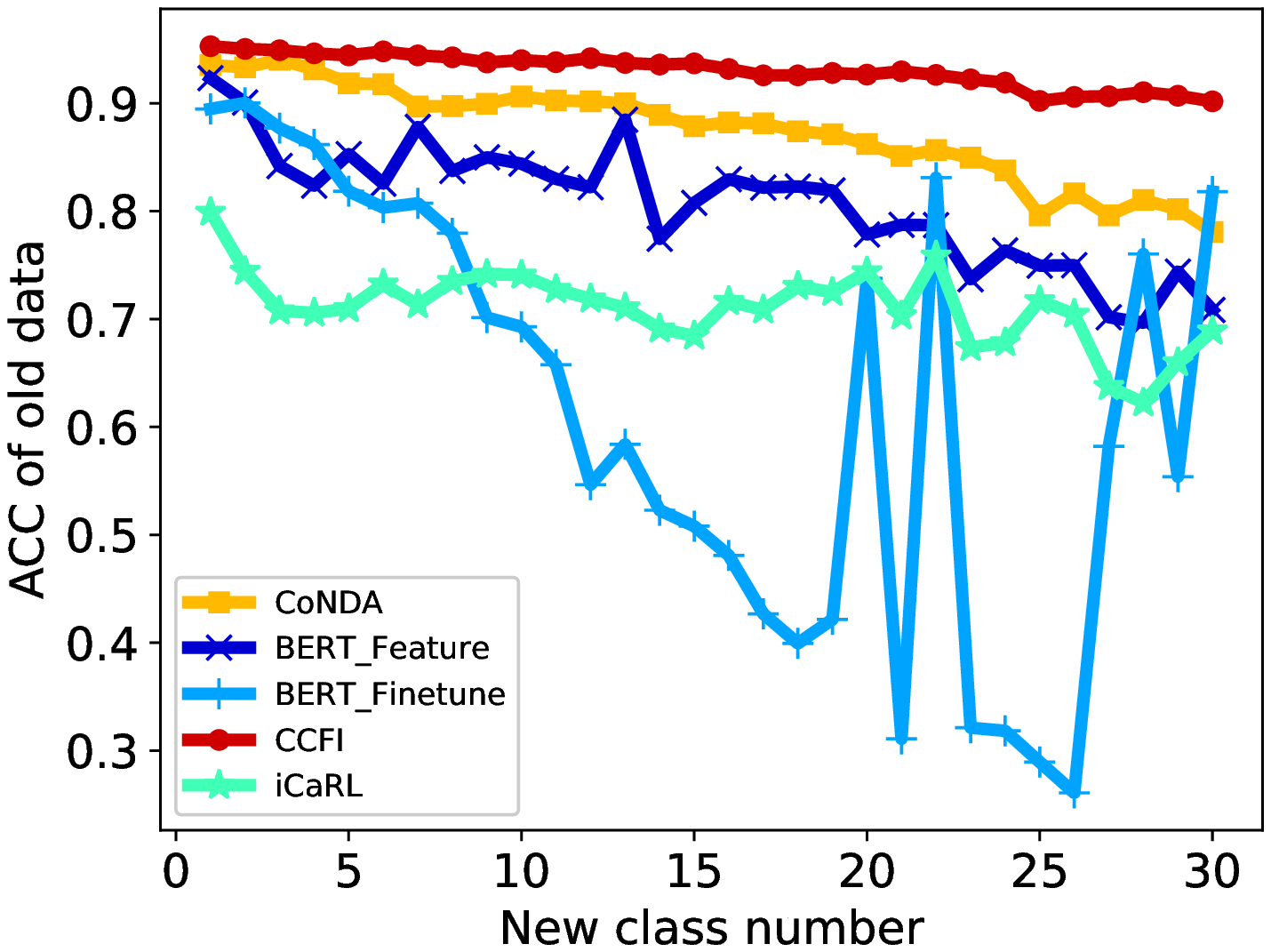}
\centering
  \caption{Old task accuracy.}
   \label{fig:cc_add_class_150:old}
\end{subfigure}
\caption{Performance of class-incremental learning on the 150-class dataset. The observable ratio of old data is kept as 10\%, 90 out of 150 classes are used for initial training, and then 30 domains are added as additional classes. }
\label{fig:cc_add_class_150}
\end{figure*}

\section{Experiment}
\label{sec:exp}
In this section, the CCFI model is first compared with the state-of-the-art methods under a continual setting. 
And further evaluations are conducted to examine the effectiveness of the individual components within CCFI model.

\subsection{Experiment Settings}
\noindent\textbf{Datasets}. We evaluated our proposed CCFI and comparison methods on  public available 150-class dataset \cite{larson-etal-2019-evaluation} and real-world (even product) 73-domain dataset 
The details of datasets can be found in Appendix \ref{app:data}.

\noindent\textbf{Baselines}. iCaRL \cite{icarl2017} and CoNDA \cite{Amazon2019continuous}, are the closest continual learning approaches to CCFI, which are designed for the scenario with access to old data. 
We also add fine-tune and the fixed feature method as baselines. 
To make fair comparisons, CCFI and all the baselines use the same BERT  backbone \cite{devlin2019bert},
and observe the same amount of old data in all learning tasks. 
The implementation details can be found in Appendix \ref{app:baseline}. 

In the main body of experiments, we report the results with the framework consisted of BERT backbone and one-layer linear classifier. 
We also conducted experiments with a multiple-layer classifier, which can be found in Appendix \ref{app:layer}.


\subsection{Quantitative Evaluation}
\label{sec:main_res}
Two key factors play in the performance of continual learning: 1) the number of new classes for retraining, and 2) the amount of old observable data. 
In this section, we first validate our model through a class-incremental learning task,  by keeping the amount of old observable data fixed and changing the number of new classes. 
We also study the effects of different exemplars by keeping the number of new classes unchanged but varying old observable data.  

\noindent\textbf{Class-incremental Learning}.
In this part,
we conduct the class-incremental learning evaluation on both 150-class and 73-domain dataset. 
Class-incremental learning can be viewed as the benchmark protocol for continual learning with access to old data \cite{icarl2017,Amazon2019continuous}. 
Specifically, after the initial training, new domains will be added in random order.
After adding each batch of new data, 
the results are evaluated on the current data set, 
considering all classes have been trained so far. 

\begin{figure*}[htbp]
  \begin{subfigure}[b]{.5\textwidth}
   \centering
\includegraphics[width=0.8\textwidth]{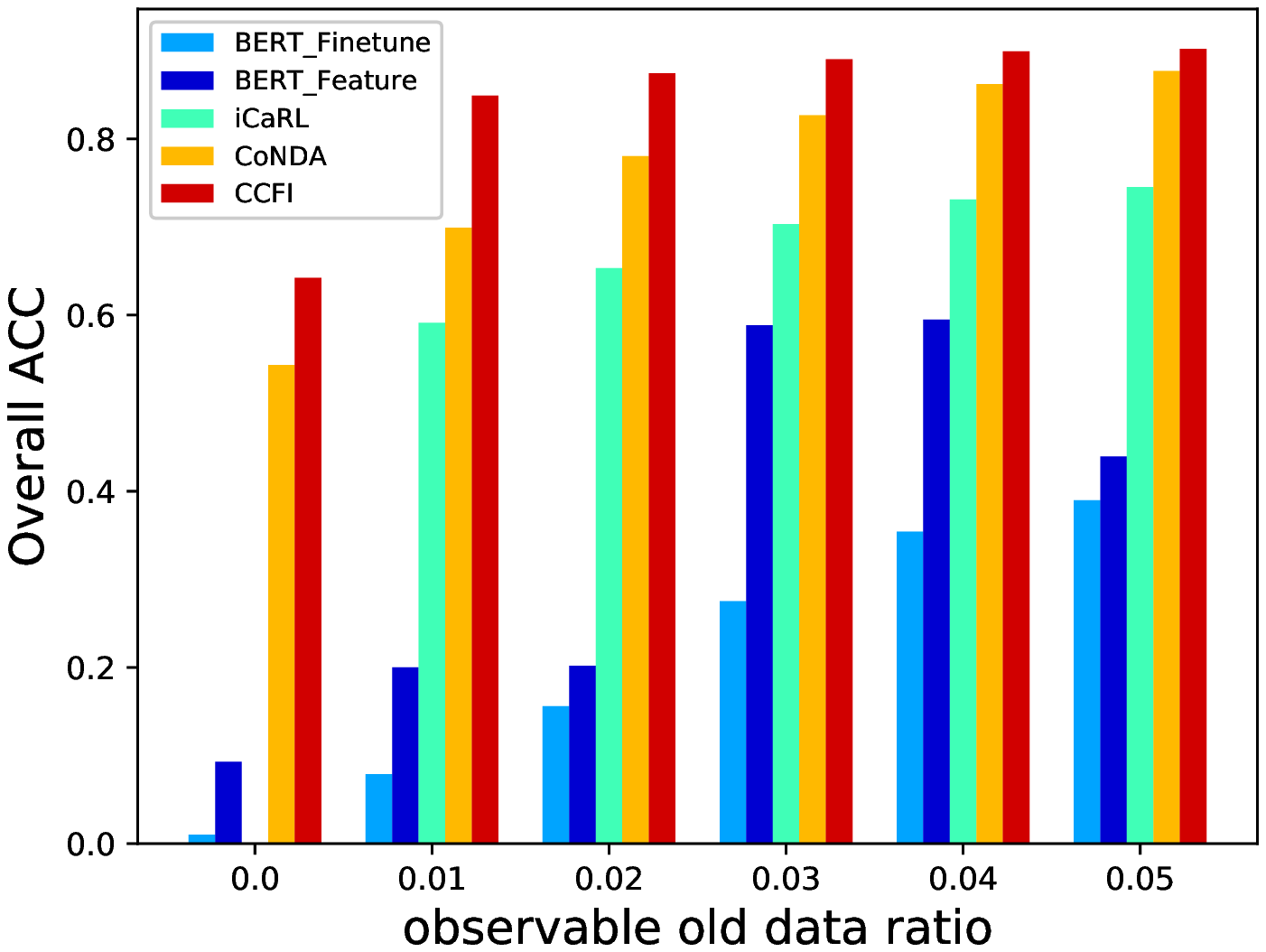}
\caption{73-domain dataset.}
\label{fig:Exemplar_73}
\end{subfigure}
  \begin{subfigure}[b]{.5\textwidth}
  \centering
\includegraphics[width=0.8\textwidth]{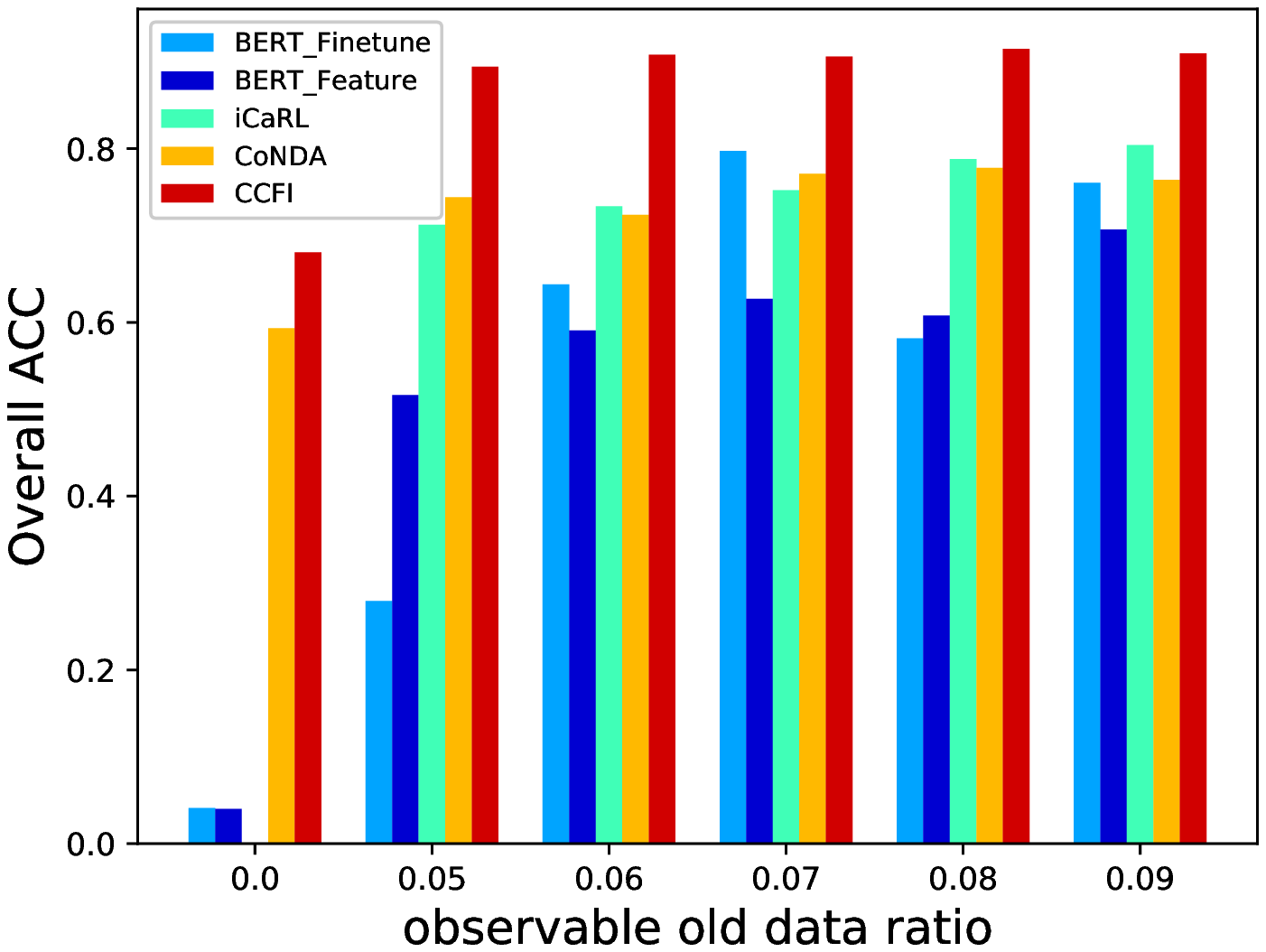}
\caption{150-class dataset.}
\label{fig:Exemplar_150}
\end{subfigure}
\caption{Effect of Differential Exemplar on the 73-domain dataset and 150-class dataset. For the 73-domain dataset, the observable ratio of old data is changed from zero to 5\%, while the number of new classes is fixed: 68 domains are used for initial training, and five domains are used as the additional new classes.
For the 150-class dataset, the observable ratio of old data is changed from zero to 10\%, while the number of new classes is fixed: 90 domains are used for initial training, and 30 domains are used as the additional new classes.
}
\label{fig:Exemplar}
\end{figure*}

Figure \ref{fig:cc_add_class_73} and Figure \ref{fig:cc_add_class_150} show the performance of class-incremental learning on 73-domain dataset and 150-class dataset. 
CCFI outperforms other methods in all tasks on both datasets.
Specifically, several observations can be made as follows.
\begin{packed_item}    \item \textbf{Overall performance}.
    Under the same new class number, CCFI always achieves the best overall accuracy among all methods. And this performance  gap is enlarged with more new classes added for retraining.
    \item \textbf{Performance fluctuations}. Fine-tune method is unstable in performance. It is the second performer on the 73-domain dataset. However, it quickly drops to almost zero and displays fluctuations on the 150-class dataset, even if the experiment conducted on the 150-class dataset is set with a higher observable ratio of old data.
    \item \textbf{Accuracy stage}. Both the fixed feature method and CoNDA display the pattern of ``performance stage'' with more new classes added,
    and CoNDA enjoys a ``larger'' stage than the fixed feature method. For example, as shown in Figure \ref{fig:cc_add_class_73:all}, 
    CoNDA maintains stable performance with 5 to 12 newly added classes varying and then suddenly drops.
    \item \textbf{Predictable performance}. 
    Both CCFI and iCaRL display linear patterns in overall performance. It means the possibility to predict and estimate class-incremental learning performance,
    which is a preferable feature in many applications. 
    But iCaRL starts at a lower accuracy and drops much faster than CCFI, probably because it discards the neural network and tunes to the simple k-nearest neighbors algorithm as the final classifier.
   This phenomenon also confirms that CCFI can enjoy the excellent performance of neural network classifiers and overcome its drawback of catastrophic forgetting. 
\end{packed_item}

\subsubsection{Different Exemplar Size}
To provide insight into the working mechanism of models capable of continual learning, 
we conduct experiments by varying the exemplar set's size with the number of new classes fixed.
Figure \ref{fig:Exemplar} shows the model performance under the effect of different exemplar sizes by changing the observable ratios of old data. 

\begin{packed_item}
    \item \textbf{Overall pattern}.
    CCFI continues to beat baselines with obvious advantages in performance. 
    Especially, CCFI can achieve high accuracy with a minimal amount (e.g.,1\%) of old data, 
    although all methods can obtain performance improvement by increasing the ratio of old observable data.
    A dramatic performance gain can also be observed from all models when the observable ratio of old data increases from zero to non-zero values. 
    This phenomenon further confirms that our experimental setting with limited access to old data is practically useful.
    \item \textbf{Consistent improvement}.
    CCFI, CoNDA, and iCarL obtain consistent improvements when increasing the ratio of old observable data. However,
    the fixed feature method doesn't get apparent benefits with more old data. 
    This phenomenon indicates more observations of old data are not the guarantee for better performance.
    And it further confirms the necessity of developing continual learning methods that can effectively utilize the information learned from exemplars.
\end{packed_item}

\subsection{Ablation Study}
\label{sec:abs}

Our proposed CCFI outperforms all the state-of-the-art methods.
To provide further insights into its working mechanism,
additional experiments are conducted to examine individual aspects of CCFI. 
\subsubsection{Dataset and Experimental Setting}
In order to avoid the occasionalities introduced by data and model complexity,
components are examined on a synthetic data set by simple neural networks with fixed weight initialization. 

Specifically, we generate a synthetic dataset of 10 completely separable classes, and each class includes 1,000 examples.
As the setting for continual learning,  we use six classes for initial training, and four classes as additional new classes for retraining. 
The neural network used in this section is a simple network with two fully-connected layers. 
The first layer is served as a feature extractor, which is fixed after the initial training.
The second layer is used as a classifier that can be tuned during retrain. 
To ensure other parts won't affect the component to be validated, the neural networks are initialized with the same weight matrix generated by a fixed random seed.
With these settings, the results can best reflect the true contribution of components.

\subsubsection{Dynamical Weight Consolidation}
\label{sec:ewc_proof}

\begin{figure*}[htbp]
  \begin{subfigure}[b]{.32\textwidth}
   \centering
   \includegraphics[width=0.88\textwidth]{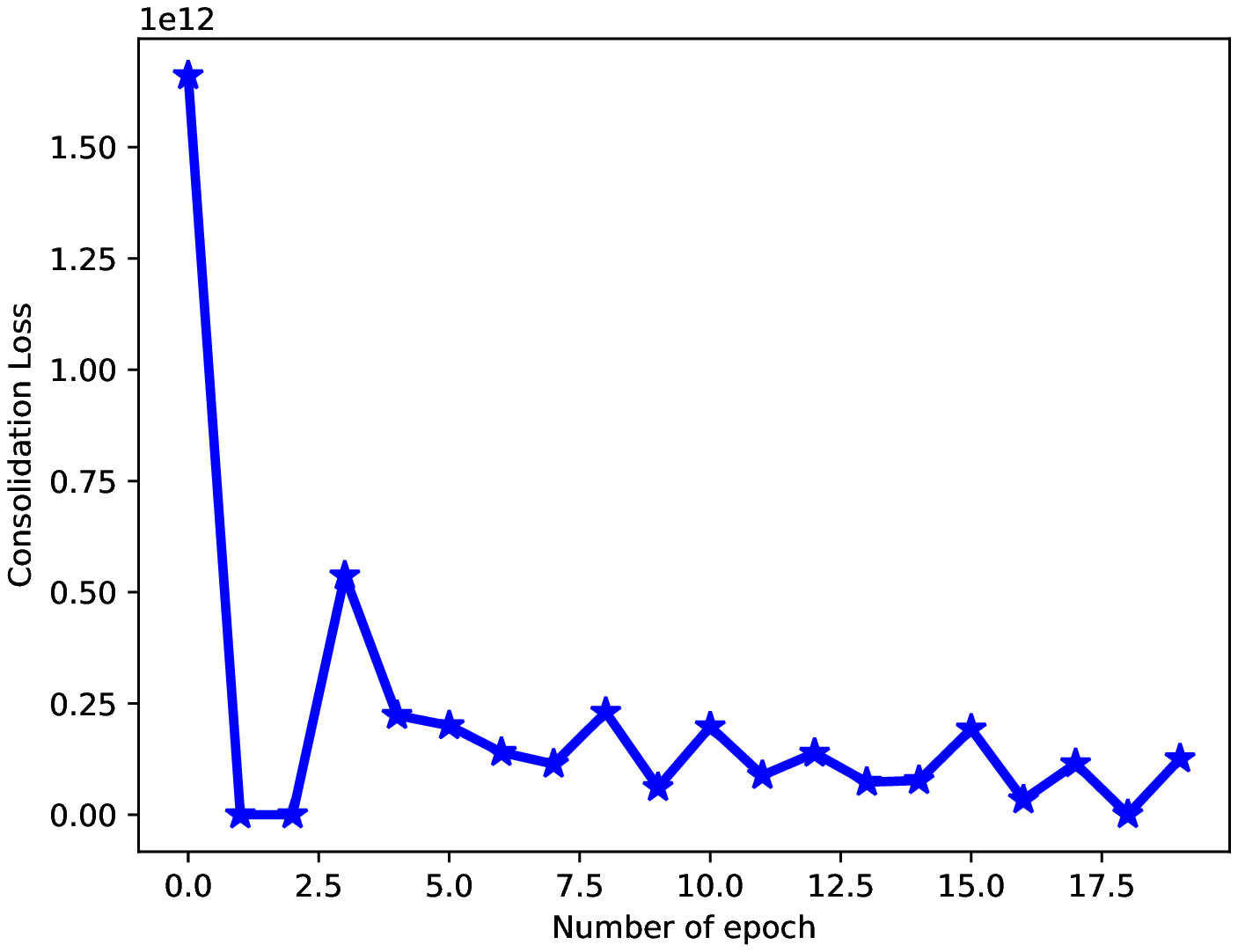}
       \caption{}
       \label{fig:ewc_loss:big}
       \end{subfigure}
\begin{subfigure}[b]{.32\textwidth}
\includegraphics[width=.88\textwidth]{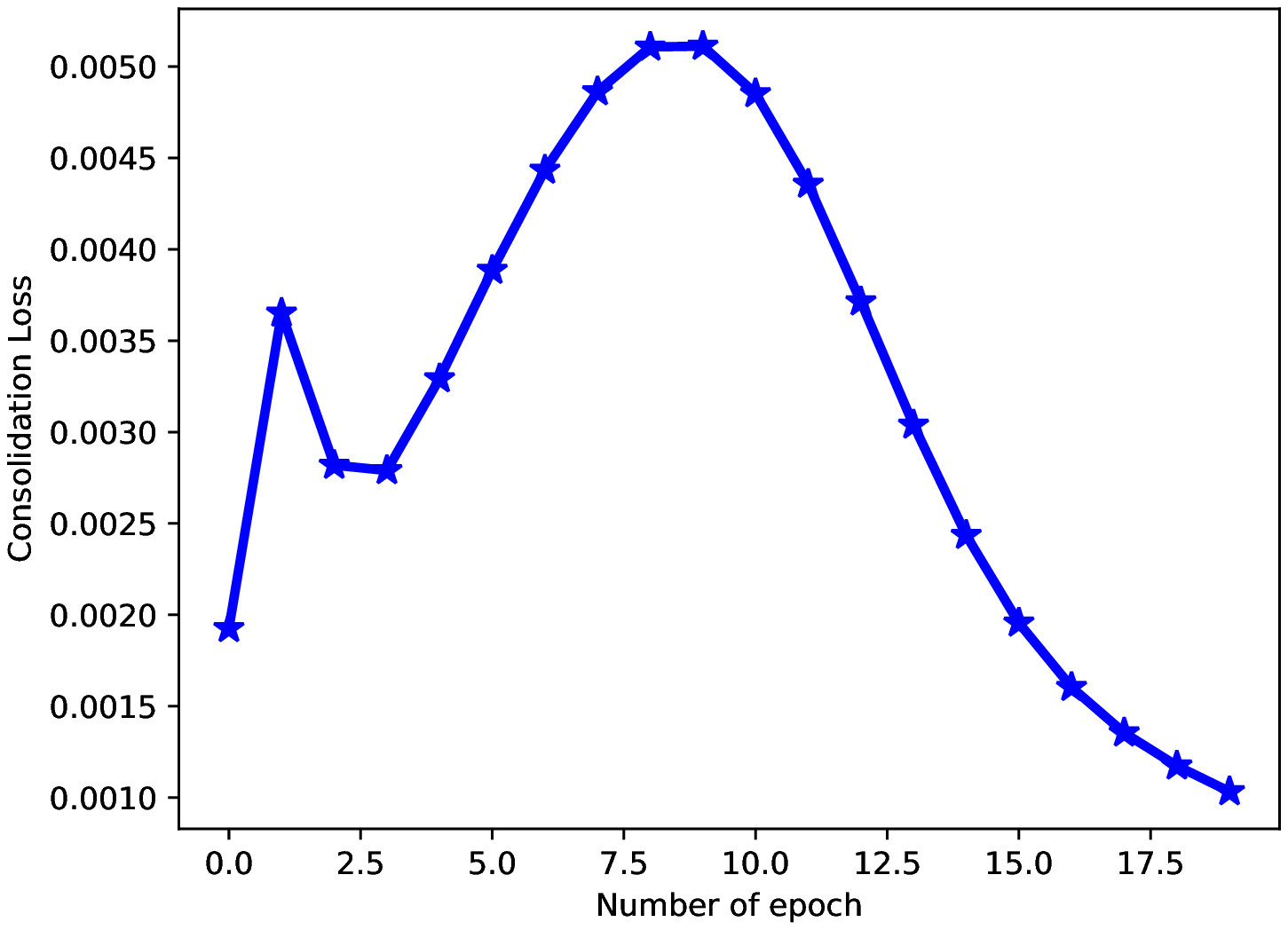}
\centering
  \caption{}
   \label{fig:ewc_loss:small}
\end{subfigure}
   \begin{subfigure}[b]{.32\textwidth}
\includegraphics[width=.88\textwidth]{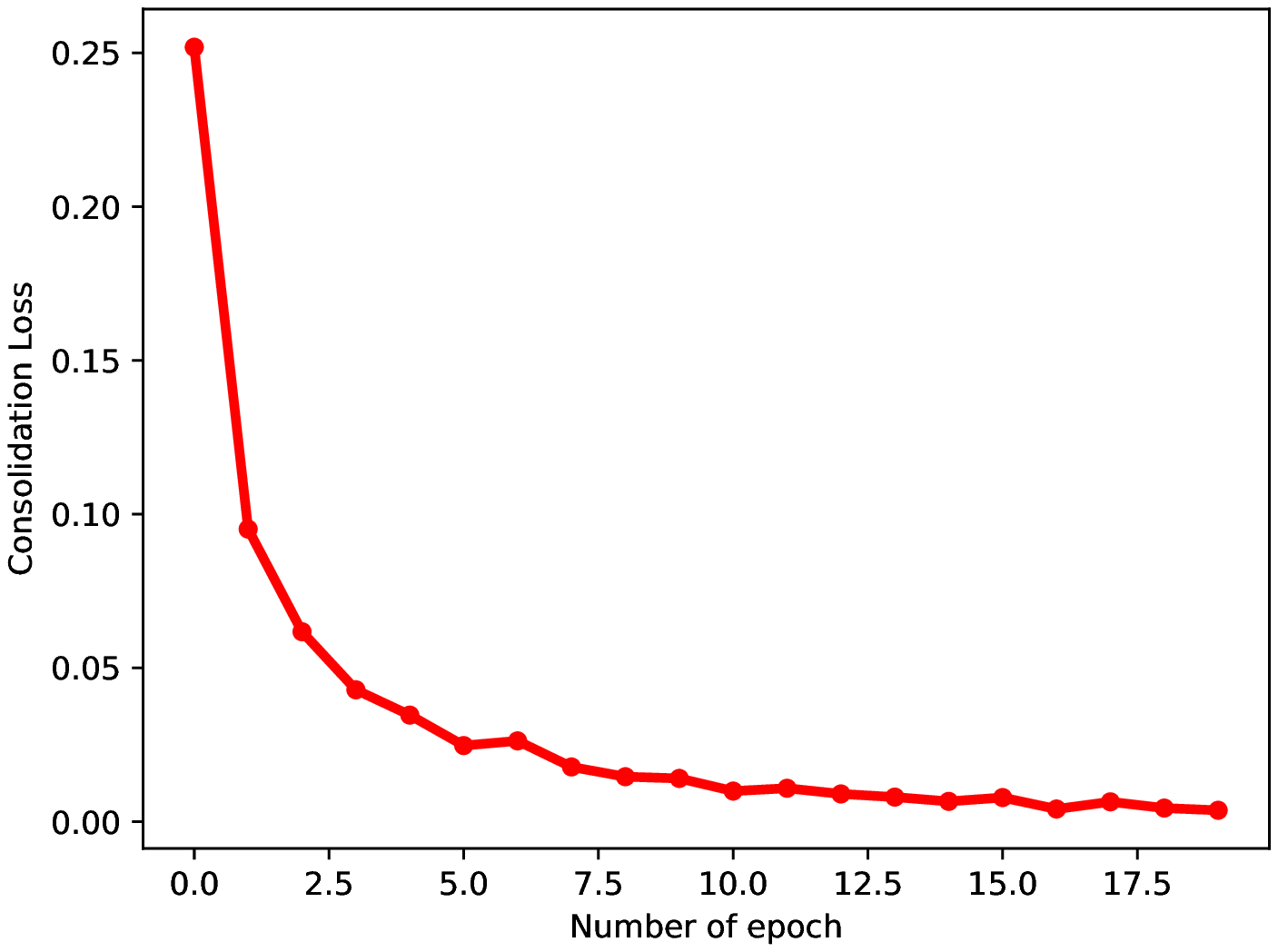}
\centering
  \caption{}
   \label{fig:ewc_loss:dynamic}
\end{subfigure}
\caption{
Consolidation loss under different weight strategies.
Figure \ref{fig:ewc_loss:big}: Fixed weight strategy  initiated by a big value $10^{25}$.
Figure \ref{fig:ewc_loss:small}: Fixed weight strategy initiated by a small value $100$. 
Figure \ref{fig:ewc_loss:dynamic}: Dynamic weight strategy using weight calculated by Equation \ref{equ:lambda}.}
\label{fig:ewc_loss}
\end{figure*}

First, we analyze the effectiveness of the dynamical weight consolidation component.
Figure \ref{fig:ewc_loss} plots the consolidation loss (second part in Equation \ref{equ:eq2}) of model using traditional fixed weight and our proposed dynamical weight consolidation.
Several observations can be made as follows.
\begin{packed_item}
    \item \textbf{Fixed weight with big value}. When the weight ($\lambda$ in Equation \ref{equ:eq2}) is set by a big value (e.g., $10^{25}$ in Figure \ref{fig:ewc_loss:big}), the consolidation loss is hard to converge and suffers from fluctuations.
    \item \textbf{Fixed weight with small value}. Oppositely, if the weight is initialized with a relatively small value (e.g., $10^{2}$ in Figure \ref{fig:ewc_loss:small}),
the consolidation loss is too small to be effective. 
In fact, as can be observed from Figure \ref{fig:ewc_loss:small}, under the small weight setting,
the consolidation loss even experiences an increase first before it slowly decreases. 
The increase in consolidation indicates that the neural network tends to sacrifice consolidation loss to lower the overall loss. 
Furthermore, this phenomenon happens when the new model modifies the important parameters learned by the original model, which are supposed to be kept with the least changes for the continual learning purpose.
\item \textbf{Dynamical weight consolidation}.
In contrast to the unstable performance of the traditional method,
as shown in Figure \ref{fig:ewc_loss:dynamic},
consolidation loss converges fast and stable by using our proposed dynamical weight consolidation.
\end{packed_item}
\begin{figure}[htbp]
  \begin{subfigure}[b]{.49\columnwidth}
   \centering
   \includegraphics[width=\columnwidth]{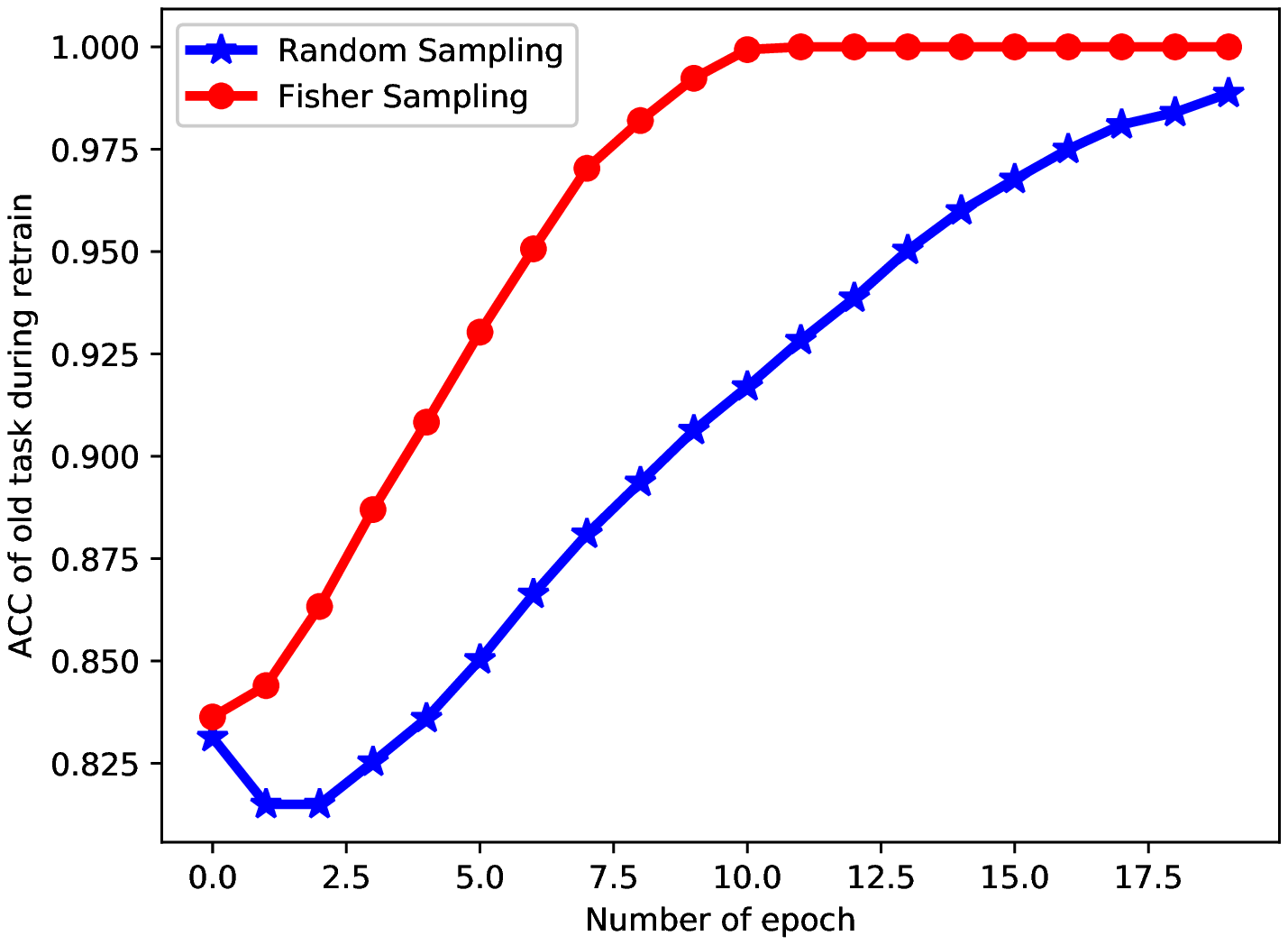}
       \caption{Sampling rate 0.5\%}
       \label{fig:sampling:0.005}
       \end{subfigure}
\begin{subfigure}[b]{.49\columnwidth}
\includegraphics[width=\columnwidth]{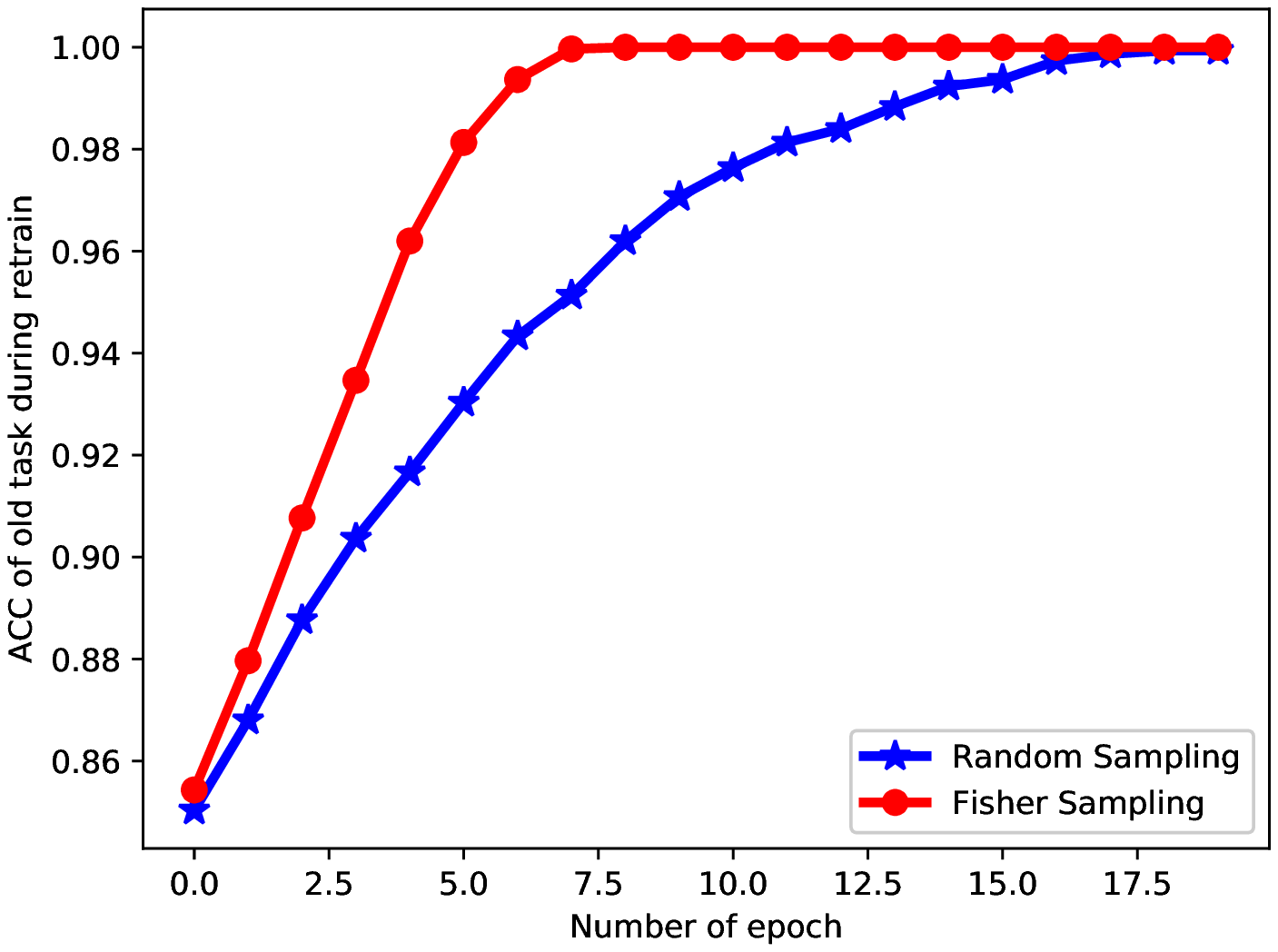}
\centering
  \caption{Sampling rate 1\%}
   \label{fig:sampling:0.01}
\end{subfigure}
   \begin{subfigure}[b]{.49\columnwidth}
\includegraphics[width=\columnwidth]{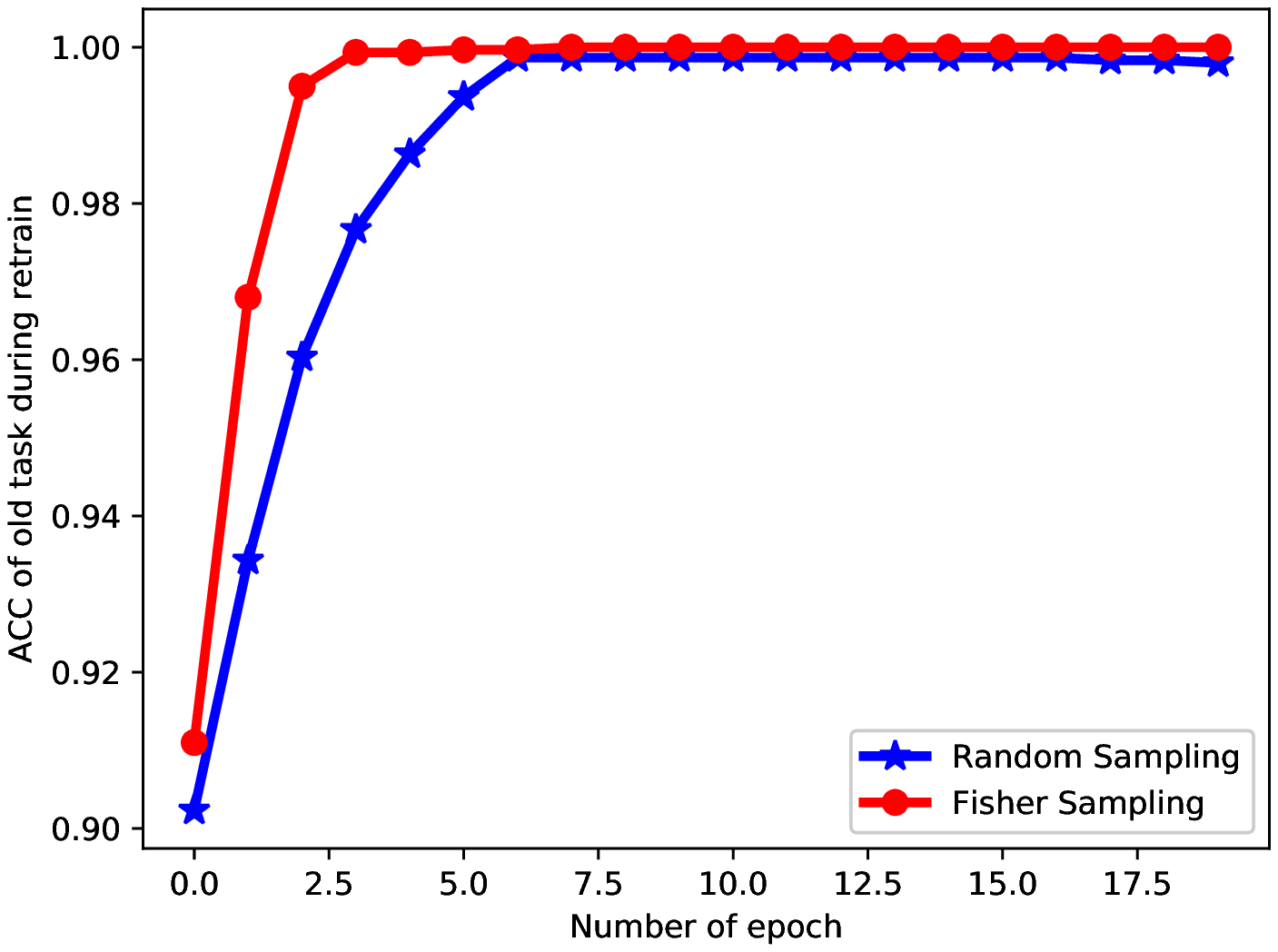}
\centering
  \caption{Sampling rate 2\%}
   \label{fig:sampling:0.02}
\end{subfigure}
   \begin{subfigure}[b]{.49\columnwidth}
\includegraphics[width=\columnwidth]{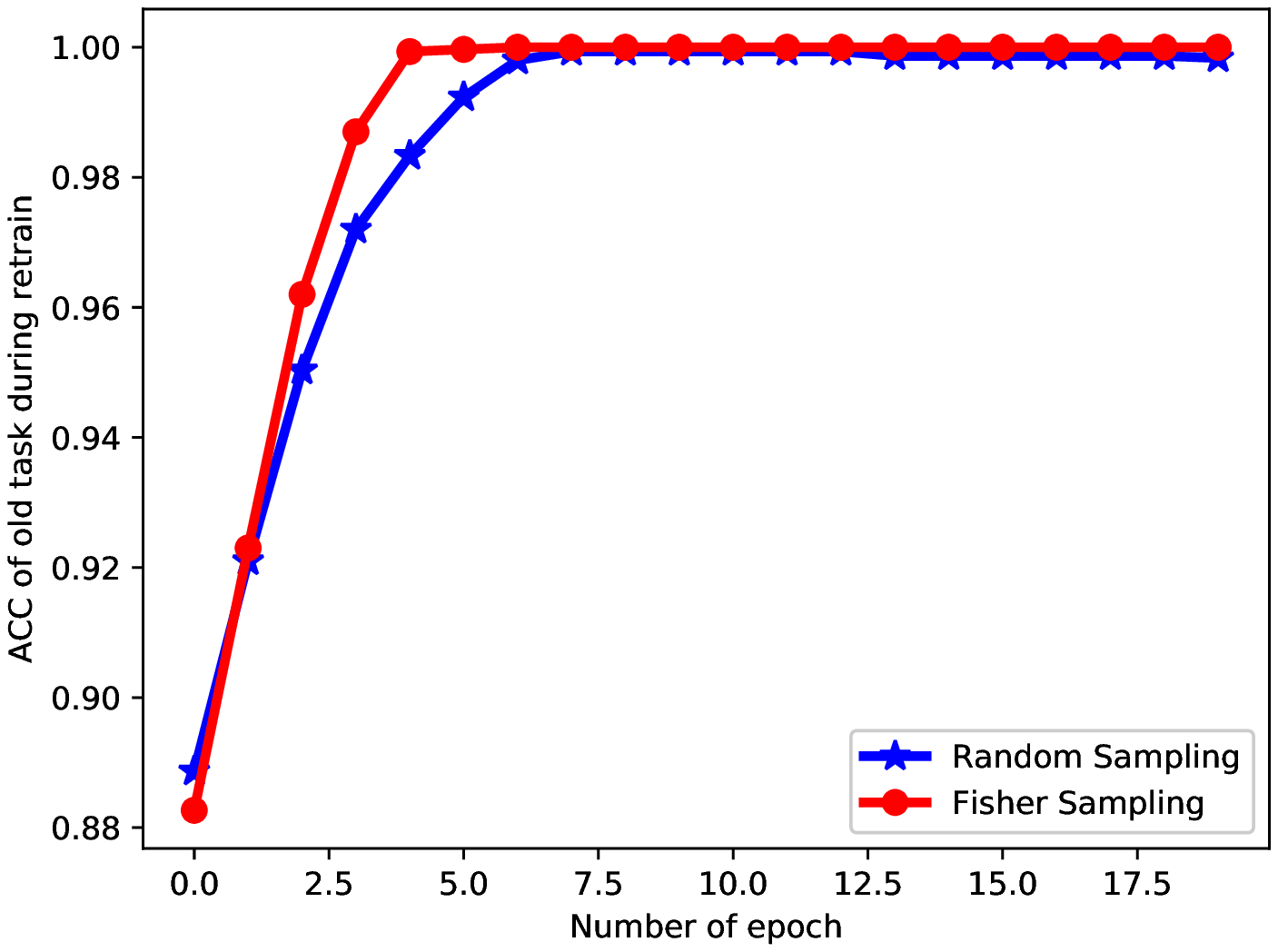}
\centering
  \caption{Sampling rate 5\%}
   \label{fig:sampling:0.04}
\end{subfigure}
\caption{
Accuracy of the old task during retraining.
Blue lines denote the accuracy using exemplars randomly selected, while red lines reflect the performance of exemplars generated by Fisher sampling. 
}
\label{fig:sampling}
\end{figure}

\subsubsection{Fisher Information Sampling}
The second set of experiments validate whether Fisher information sampling is indeed beneficial to the overall performance, compared with using randomly selected examples.

To examine how much improvement can be obtained by Fisher sampling alone, we remove the weight consolidation component in this section.
Thus the results reported here are outputs of the simple two-layer model by using exemplars during retraining. 
From another point of view, these results show the amount of information the exemplars carrying from the original model.

Figure \ref{fig:sampling} plots the accuracy of the old task during the retraining process.
Although the network is retrained with only a small set of old data, the accuracy is computed over all old data to fully examine the quality of exemplars.
Since the classes in synthetic data are fully separable, the accuracy will be 100\% eventually.
Therefore, the quality of exemplars is demonstrated by the converging speed.
A faster converge speed provided by an exemplar set is of great significance in three aspects:
\begin{packed_item}
    \item \textbf{Better computational efficiency}. With the same amount of old data for retraining, the most obvious benefit indicated by the faster converge speed is, the better computational efficiency since the model requires less retraining time.
    \item \textbf{Less ``damage'' to original model}.
    A faster converge speed indicates less ``damage'' to the original model. 
    All weight consolidation schemes act like ``buffers'' for the old parameter.
    With these schemes, old parameters will slow down their changes when new tasks come. 
    To best cope with the consolidation schemes,
    good exemplars should achieve comparable good performance with fewer retraining epochs,
    since more retraining epochs mean that the new model has modified more parameters from the original network. 
    \item \textbf{More information of original dataset}.
    As mentioned above, under the synthetic data is fully separable, the accuracy will be 100\% eventually.
    In this case, a faster speed can be ``converted'' to more information, as experiments with more data always require fewer epochs to reach the states of convergence.
    For example, as shown in Figure \ref{fig:sampling},  much more epochs are needed under sampling rate 0.5\% than that of 1\%.
    
\end{packed_item}
 Figure \ref{fig:sampling} shows,
exemplars generated by Fisher sampling can achieve much faster converge speed than randomly selected exemplars,
which  proves Fisher sampling alone can contribute contribution effectively to the overall performance.

\section{Conclusion}
\label{sec:con}
This paper proposes  a hyperparameter-free model called CCFI  for continuous domain classification.
 CCFI can record information of old models via exemplars selected by Fisher information sampling, and conduct efficient retraining through dynamical weight consolidation.
The comparison against the existing models reveals CCFI is the best performer under various experimental environments, without additional efforts in hyperparameter searching.

\clearpage

\bibliography{emnlp2020,custom}
\bibliographystyle{acl_natbib}
\appendix

\begin{table*}[htbp]
\centering
\small
\begin{tabular}{ccc|cc|cc|cc}\toprule
\multicolumn{1}{c|}{}& \multicolumn{2}{c|}{ Finetune } & \multicolumn{2}{c|}{ Feature Extraction} & \multicolumn{2}{c|}{ CoNDA } & \multicolumn{2}{c}{ CCFI } \\
\cline{2-9}
\multicolumn{1}{c|}{classifier}& Initial training & Retrain & Initial training & Retrain & Initial training & Retrain & Initial training & Retrain \\
\midrule
\multicolumn{1}{c|}{One-layer} & 0.9671 & 0.7232 & 0.9671 & 0.8556 & 0.9566 & 0.9116 & 0.9614 & 0.9426 \\
\multicolumn{1}{c|}{Two-layer} & 0.9592 & 0.2387 & 0.9592 & 0.6495 & 0.9522 & 0.9366 & 0.9541 & 0.9408 \\
\bottomrule
\caption{Effect of classifier layer number on the 150-class dataset.
90 out of 150 classes are used for initial training, and then additional data of 10 domains plus 10\% of old data (90-class) are used for retraining.}
\label{tab:layers-150}
\end{tabular}
\end{table*}

\begin{table*}[htbp]
\centering
\small
\begin{tabular}{ccc|cc|cc|cc}\toprule
\multicolumn{1}{c|}{}& \multicolumn{2}{c|}{ Finetune } & \multicolumn{2}{c|}{ Feature Extraction} & \multicolumn{2}{c|}{ CoNDA } & \multicolumn{2}{c}{ CCFI } \\
\cline{2-9}
\multicolumn{1}{c|}{classifier}& Initial training & Retrain & Initial training & Retrain & Initial training & Retrain & Initial training & Retrain \\
\midrule
\multicolumn{1}{c|}{One-layer} & 0.9561 & 0.6322 & 0.9561 & 0.2839 & 0.9349 & 0.4375 & 0.9587 & 0.7378 \\
\multicolumn{1}{c|}{Two-layer} & 0.9422 & 0.0751 & 0.9422 & 0.1193 & 0.9314 & 0.6868 & 0.9472 & 0.7134 \\
\bottomrule
\caption{Effect of classifier layer number on the 73-domain dataset.
43 out of 73 classes are used for initial training, and then additional data of 10 domains plus 1\% of old data (43-class) are used for retraining.}
\label{tab:layers-73}
\end{tabular}
\end{table*}

\section{Appendix}

\subsection{Dataset Statistics}
\label{app:data}

The general statistics of the 150-class dataset \footnote{https://github.com/clinc/oos-eval} and 73-domain dataset are listed below. 
\begin{packed_item}
    \item 150-class dataset: balanced dataset with 150 intents that can be grouped into 10 general domains. Each intent has 100 training queries, 20 validation, and 30 testing queries.
    \item 73-domain dataset: imbalanced dataset with 73 domains.
    Each domain contains 512 examples on average.
    However, this dataset is highly imbalanced that the largest domain includes 1,771 examples, while the smallest domain only has 27 examples. 
\end{packed_item}
In both datasets, we split examples of each class into 90\% for training, 5\% for validation, and 5\% for testing.
All classification accuracy results are evaluated on the test set.
\subsection{Implement Details}
\label{app:baseline}
\textbf{Specific settings}. In our implement of CoNDA, we pick up hyperprameter $\Delta_{pos}=0.5$ and $\Delta_{neg}=0.3$.
The fixed-feature method freezes 12 layers of BERT after the initial training. Only the parameters in the new classifier layer are allowed for tuning, which in a way provides the function of continual learning. 
Fine-tune method can modify parameters in all 12 layers of BERT, which can be viewed as the network with little prevention of catastrophic forgetting.

\noindent\textbf{General settings}. Adam optimizer is used in all learning processes, and the learning rate is always set to be $0.00001$. 
All runs are trained on 4 V100 GPUs with a batch size of 32.
Our example code can be found at:
\url{https://github.com/tinghua-code/CCFI}

\subsection{Multi-layer Classifier Results}
\label{app:layer}
To examine the effect of classifier layer number (amount of retrainable parameters), we run experiments under two frameworks.
The first framework is the same as the one used in the main experimental part, which consists of a 12-layer BERT feature extractor and a one-layer linear classifier.
The second framework keeps the BERT feature extractor unchanged and adds one more layer to the classifier.
The results are listed in Table  \ref{tab:layers-150} and \ref{tab:layers-73}, and several observations can be made as follows.
\begin{packed_item}
    \item CCFI still remains the best performer. Our proposed CCFI model produces good performance regardless of the number of layers in the classifier. This phenomenon further confirms its effectiveness and stability.  
    \item CoNDA is the second-best performer in both frameworks. Notably, the retraining performance of CoNDA increases when we increase the number of layers.
    \item BERT finetune and feature extraction method become worse when increasing the number of layers. These two baselines are sensitive to the structure of the classifier, which indicates the superficial variations of pre-trained models are not enough for continual learning.
    \item One-layer classifier works well enough with BERT.  As can be seen from Table \ref{tab:layers-150} and \ref{tab:layers-73}, the initial training results of all methods degrade when increasing the number of classifier layers. Therefore, we report the results based on a one-layer linear classifier in the main body of the paper. 
    
\end{packed_item}

\end{document}